\documentclass[sigconf]{acmart}

\AtBeginDocument{%
  }
    
\copyrightyear{2023}
\acmYear{2023}
\setcopyright{acmlicensed}\acmConference[MM '23]{Proceedings of the 31st ACM International Conference on Multimedia}{October 29-November 3, 2023}{Ottawa, ON, Canada}
\acmBooktitle{Proceedings of the 31st ACM International Conference on Multimedia (MM '23), October 29-November 3, 2023, Ottawa, ON, Canada}
\acmPrice{15.00}
\acmDOI{10.1145/3581783.3612395}
\acmISBN{979-8-4007-0108-5/23/10}

\usepackage{multirow}

\newcommand{\blue}[1]{\textcolor{blue}{#1}}
\newcommand{\ie}{\textit{i}.\textit{e}., }
\newcommand{\eg}{\textit{e}.\textit{g}., }

\begin{document}

\title{Do Vision-Language Transformers Exhibit Visual Commonsense? An Empirical Study of VCR}

\author{Zhenyang Li}
\affiliation{%
  \institution{Shandong University}  
  \streetaddress{72 Binhai Highway, Qingdao, Shandong Province, P.R. China}   
  \city{}     
  \country{}
  }
\email{zhenyanglidz@gmail.com}

\author{Yangyang Guo}
\affiliation{%
  \institution{National University of Singapore}  
  \streetaddress{11 Research Link}
  \city{}
  \country{}
  }
\email{guoyang.eric@gmail.com}

\author{Kejie Wang}
\affiliation{%
  \institution{Shandong University}  
  \streetaddress{72 Binhai Highway}
  \city{}
  \country{}
  }
\email{kjwang.henry@gmail.com}

\author{Xiaolin Chen}
\affiliation{%
  \institution{Shandong University}  
  \streetaddress{72 Binhai Highway}
  \city{}
  \country{}
  }
\email{cxlicd@gmail.com}

\author{Liqiang Nie}
\authornote{Corresponding Author: Liqiang Nie.}
\affiliation{%
  \institution{Harbin Institute of Technology, Shenzhen}  
  \streetaddress{HIT Campus of University Town of Shenzhen}
  \city{}
  \country{}
  }
\email{nieliqiang@gmail.com}

\author{Mohan Kankanhalli}
\affiliation{%
  \institution{National University of Singapore}  
  \streetaddress{11 Research Link}
  \city{}
  \country{}
  }
\email{mohan@comp.nus.edu.sg}
\renewcommand{\shortauthors}{Zhenyang Li et al.}

\begin{abstract}
Visual Commonsense Reasoning (VCR) calls for explanatory reasoning behind question answering over visual scenes. 
To achieve this goal, a model is required to provide an acceptable rationale as the reason for the predicted answers.
Progress on the benchmark dataset stems largely from the recent advancement of Vision-Language Transformers (VL Transformers). 
These models are first pre-trained on some generic large-scale vision-text datasets, and then the learned representations are transferred to the downstream VCR task.
Despite their attractive performance, this paper posits that the VL Transformers do not exhibit visual commonsense, which is the key to VCR.
In particular, our empirical results pinpoint several shortcomings of existing VL Transformers: small gains from pre-training, unexpected language bias, limited model architecture for the two inseparable sub-tasks, and neglect of the important object-tag correlation.
With these findings, we tentatively suggest some future directions from the aspect of dataset, evaluation metric, and training tricks.
We believe this work could make researchers revisit the intuition and goals of VCR, and thus help tackle the remaining challenges in visual reasoning.
\end{abstract}



\maketitle

\section{Introduction}
\begin{figure}[!t]
  \centering
  \includegraphics[width=1.0\linewidth]{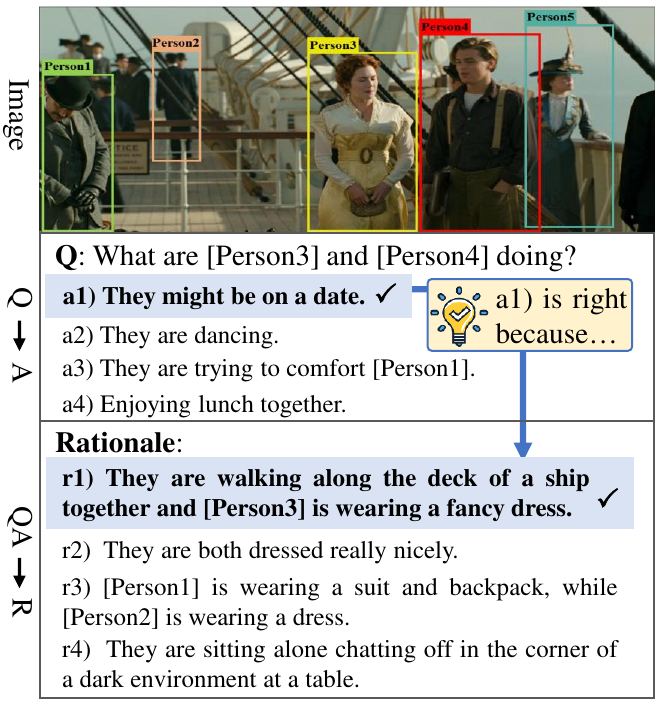}
  \caption{An exemplar of VCR. The task is composed of two sub-tasks: Q$\rightarrow$A and QA$\rightarrow$R, where the challenge mainly lies in the cross-modal reasoning from the latter.}\label{fig:teaser}
\end{figure}

The intersection of vision and language areas has spawned numerous multi-modal tasks, such as image captioning~\cite{image_caption1,image_caption2,image_caption3} and Visual Question Answering (VQA)~\cite{vqa1,vqa2, multimodal_relation_VQA2} over the past few years. Among these, Visual Commonsense Reasoning (VCR)~\cite{r2c} has recently drawn increasing attention from researchers due to its challenging nature. Beyond answering visual questions as conventional VQA does (Q$\rightarrow$A), VCR further requires the model to pick the rationale for the Q$\rightarrow$A process (where the visual commonsense is), namely QA$\rightarrow$R (see Figure~\ref{fig:teaser}).

\begin{figure*}[!t]
    \centering
    \includegraphics[width=1.0\linewidth]{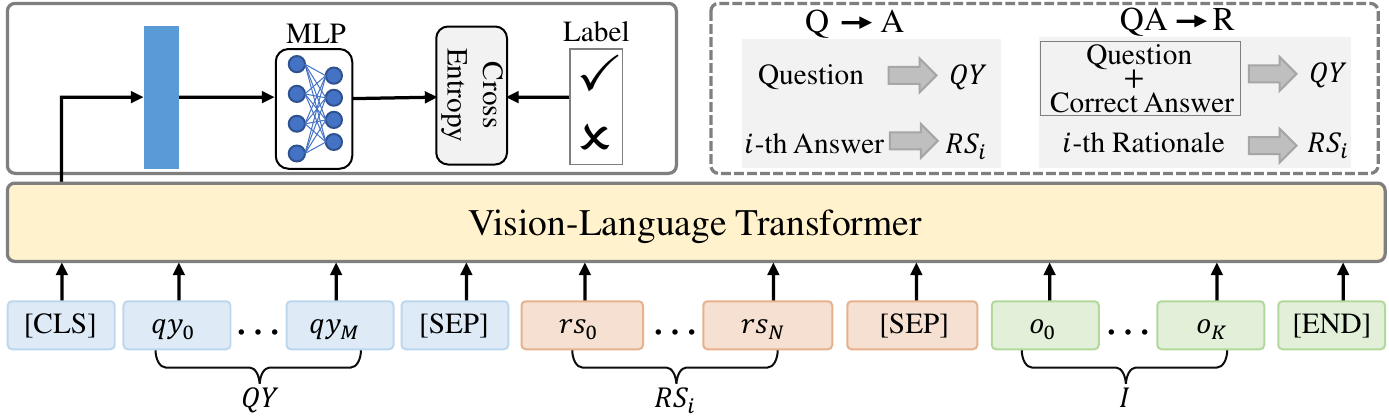}
    \caption{Pipeline of Vision-Language Transformers for VCR.
    Q$\rightarrow$A and QA$\rightarrow$R share the same pipeline where only the input query ($QY$) and response ($RS$) are slightly different.} 
    \label{fig:pipeline}
\end{figure*}

VCR is taken as an important proxy for visual commonsense understanding.
To deal with this difficult task, some initial efforts have been devoted to designing task-specific model architectures~\cite{r2c,CCN, R3_MM}. 
These models utilize the contextualized query-region affinity captured by well-designed attention mechanisms as evidence, to reason the plausibility between responses and images.
Subsequently, Vision-Language Transformers (VL Transformers)~\cite{VL-BERT,vilbert, KVL-BERT} swept the multi-modal vision-language domain and rapidly prevailed over the competing counterparts on the VCR Leaderboard\footnote{https://visualcommonsense.com/leaderboard/.}.
VL Transformers first pre-train BERT style models on generic vision-language datasets (such as Conceptual Captions~\cite{cc})  for task-agnostic representation learning, which is then transferred to the downstream VCR for both Q$\rightarrow$A and QA$\rightarrow$R.

Though the state-of-the-art keeps advancing, the reasoning capability of these VL Transformers still remains debatable. As an improvement of VQA, VCR is promising not simply because the performance of traditional VQA benchmarks has saturated, but it is expected to uncover the complex reasoning behind answer prediction, \ie rationale prediction~\cite{r2c}.
On the flip side, the key to the success of VL Transformers, namely pretext training objectives (\eg masked language modeling), deviates substantially from the reasoning goal. 
In particular, typical pretext tasks usually focus either on the reconstruction from partially masked elements, or the coherence between the two given modalities. However, why these modality matching-driven objectives aid visual reasoning on VCR remains less persuasive.

\begin{figure*}[!t] 
  \centering
 \includegraphics[width=1.0\linewidth]{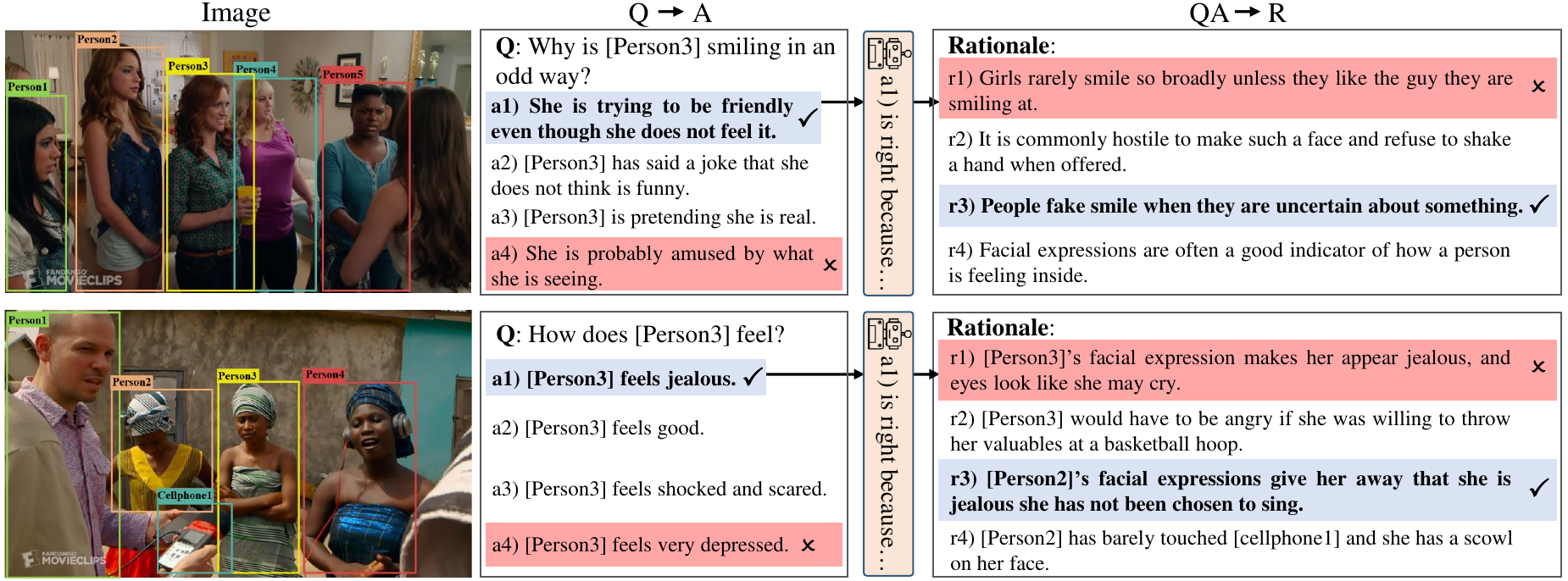}
  \caption{Failure cases from VILLA. The input to QA$\rightarrow$R consists of the correct answer (blue one from Q$\rightarrow$A), rather than the predicted answer (red one from Q$\rightarrow$A)  following the default setting. It can be seen that the model makes mistakes on cases calling for fine-grained reasoning.}
  \label{fig:case_study}
\end{figure*}

Given the above concern, in this work, we empirically find that VL Transformers perform well mostly in those instances requiring less reasoning while failing on difficult ones (refer to Figure~\ref{fig:case_study}). We then conduct an in-depth investigation into this problem and obtain the following findings:

\begin{itemize}
    \item Limited benefits are transferred from pre-training to VCR. Pre-training on large-scale vision-language datasets enhances some downstream tasks like image retrieval with significant performance margin~\cite{uniter,vilbert}. 
    In contrast, VCR improves little from these carefully designed pre-training steps. 
    We attribute this finding to two reasons: 1) domain shift between pre-training and VCR fine-tuning, and 2) weak reasoning of these pretext objectives. 
    
    \item Language bias prevents the model from cross-modal reasoning. 
    The language shortcut between textual queries and responses leads the model to make decisions based on the text modality only~\cite{shortcut_in_vcr}, especially for QA$\rightarrow$R.
    When it comes to cases that require visual reasoning, the model is misled to leverage the bias between text due to their overwhelming co-occurrence than that of image and text. 
    
    \item The architecture does not lend itself to a holistic solution for both Q$\rightarrow$A and QA$\rightarrow$R. 
    Based on the definition and intuition of VCR, the Q$\rightarrow$A and QA$\rightarrow$R should be made  consistent rather than being treated separately~\cite{tip}. 
    Unfortunately, existing VL Transformer models are limited in handling these two sub-tasks with consistency. 
    
    \item The unique tag labels are somehow under-utilized by current VL Transformers. 
    As demonstrated in Figure~\ref{fig:teaser}, the tag (such as `[person3]') defines an exclusive link between an object label and a certain bounding box.
    It is essential to consider such relationships for proper reasoning rather than treating them as being independent.
\end{itemize}

This paper shows that the above drawbacks are common among some representative VL Transformers. 
Nevertheless, it is not our goal to propose a novel method to close the gap. 
Instead, the key contribution of this work lies in its insights for developing new methods, which helps bypass certain limitations.
Towards this, we also outline several potential research directions given the analysis of these problems. 
\section{Preliminary}

\subsection{Problem Formulation}
Given a natural image and a textual question, Visual Commonsense Reasoning (VCR) aims to predict the answer to this question as well as the explanation of the answering process. 
Compared to VQA, the questions in VCR are made more challenging and the models are expected to provide the rationale behind the question answering. 
Accordingly, VCR focuses mainly on higher-order cognition and commonsense understanding of images. 
Specifically, a typical VCR model can be formulated as follows,
\begin{equation}
    RS = \mathop{\mathrm{argmax}}\limits_{{RS}_i\in{\mathcal{RS}}} {f(I, QY, {RS}_i | \mathbf{\Theta} )},
\end{equation}
where $I$ and $QY$ are the given image and query, respectively; 
$\mathcal{RS}$ denotes the response set where ${RS}_i$ is the $i$-th element, $\mathbf{\Theta}$ denotes the involved optimized parameters, and the function $f$ predicts a compatible score based on the given inputs. 
In practice, VCR is often decomposed into the following two multiple-choice sub-tasks:

\noindent \textbf{Question answering (Q$\rightarrow$A)} -- For the given image $I$ and a corresponding question $Q \Leftarrow QY$, the model is required to choose the right answer $A \Leftarrow RS$ from a set of answer choices $\mathcal{A} \Leftarrow \mathcal{RS}$.

\noindent \textbf{Answer justification (QA$\rightarrow$R)} -- Similar to the inputs of Q$\rightarrow$A, the model in this stage takes the correct answer $A$ as the additional input ($Q+A \Leftarrow QY$), and is expected to select the right rationale $R \Leftarrow RS$ from a set of rationale choices $R \Leftarrow \mathcal{RS}$.

\subsection{VL Transformer Pre-training} \label{sec:pre-train-objective}
The past few years have witnessed the rapid development of VL Transformers. 
In addition to the large-scale datasets, the pretext pre-training tasks or objectives are the key to the success of these models. 
We summarize three typical widely applied tasks, \ie Cross-modal Masked Language Modeling (MLM), Masked Region Classification (MRC), and Image-Text Matching (ITM). 

\noindent \textbf{MLM} originates from the MLM task in BERT~\cite{bert}. 
The key difference is that the visual clues are incorporated in VL Transformers for capturing the dependencies between linguistic and visual contents,
\begin{equation}
    \mathcal{L}_{MLM} = -\mathbb{E}_{(T,I)\in D}\log P_{\mathbf{\theta}}(t_m|T_{\setminus m}, I),
\end{equation}
where $\mathbf{\theta}$ represents the parameters of the VL Transformer, $t_m$ and $T_{\setminus m}$ denote the masked and the remaining tokens, respectively. 
Each pair $(T, I)\in D$ is composed of a text $T$ and an image $I$ sampled from a vision-language dataset $D$.

\noindent \textbf{MRC} is a dual task of MLM. 
It learns to predict the semantic class of each masked object based on the corresponding text and its surrounding visual objects. 
To pre-train this, the cross-entropy loss ($CE$) between the output distribution normalized by a softmax function $s(i_n^{'})$ and class label $c(i_n)$ for the masked region $i_n$ is employed,
\begin{equation}
    \mathcal{L}_{MRC} = \mathbb{E}_{(T,I)\in D}\sum_{n=1}^{N}CE(s(i_n^{'}), c(i_n)),
\end{equation}
where $s(i_n^{'})$ denotes the VL Transformer output of $i_n$ and $N$ is the number of the objects detected from image $I$.

\noindent \textbf{ITM} is similar to the Next Sentence Prediction task utilized in BERT~\cite{bert}. 
Given an image-text pair as input, the Transformer must predict whether the image and text are aligned, \eg whether the text describes the image,
\begin{equation}
\begin{split}
    \mathcal{L}_{ITM} = -\mathbb{E}_{(T,I)\in D}[&y\log s_{\theta}(T, I)\\
    + &(1-y)\log (1 - s_{\theta}(T, I))],
\end{split}
\end{equation}
where $y$ is the ground truth and $s_{\theta}$ is the score function to measure the alignment probability of $(T,I)$.

\subsection{Fine-tuning on VCR}

\noindent \textbf{Input Formats} -- Figure~\ref{fig:pipeline} illustrates the input format of typical VL Transformers in VCR. 
Like in other VL tasks, the inputs are composed of a special classification token [CLS], the corresponding text, a separation token [SEP] between the two modalities, the given image, and an end token [END]. 
Specifically, pertaining to the textual input, the concatenation of $Q$ and $A_i$ with a [SEP] is applied for the Q$\rightarrow$A sub-task;
while for QA$\rightarrow$R, the usual way is to concatenate $Q$, ground-truth answer $A$ and the candidate rationale $R_i$. 

\begin{table*}[!ht]
    \centering
    \caption{Pre-training gains from VILLA on five cross-modality tasks.}
    \scalebox{1.0}{
    \begin{tabular}{c|*{8}{c}}
    \toprule
    \multirow{2}{*}{Pre-train} & \multicolumn{2}{c}{$\text{NLVR}^2$}    & \multicolumn{2}{c}{Retrieval}   & \multirow{2}{*}{VQA}      & \multicolumn{3}{c}{VCR Validation}     \\
                                \cmidrule(lr){2-3}                      \cmidrule(lr){4-5}                                                   \cmidrule(lr){7-9}
                                & dev       & test-P                     & Text & Image                   &                           & Q$\rightarrow$A   & QA$\rightarrow$R  & Q$\rightarrow$AR  \\ 
    \midrule
         $\times$              & 50.9      & 51.2                       & 80.5 & 65.4                     & 68.4                      & 72.6              & 75.1              & 54.7              \\
         \checkmark            & 78.4      & 79.3                       & 86.6 & 74.7                     & 73.6                      & 73.9              & 76.1              & 56.5              \\
    \midrule
         $\Delta$              & 27.5      & 28.1                       & 6.1  & 9.3                      & 5.2                       & \blue{+1.3}       & \blue{+1.0}       & \blue{+1.8}        \\
    \bottomrule
    \end{tabular}}
    \label{tab:villa-pretrain}
\end{table*}

\begin{table}[t]
    \centering
    \caption{Pre-training gains from UNITER on three tasks.}
    \scalebox{1.0}{
    \begin{tabular}{c|*{5}{c}}
    \toprule
         \multirow{2}{*}{Pre-train} & \multicolumn{2}{c}{Retrieval}  & \multicolumn{3}{c}{VCR Validation}                        \\
                                    \cmidrule{2-3}                     \cmidrule(lr){4-6}
                                    & Text          & Image          & Q$\rightarrow$A   & QA$\rightarrow$R  & Q$\rightarrow$AR  \\ 
    \midrule
         $\times$                   & 83.3          & 73.9           & 71.5              & 72.9              & 52.2                 \\
         \checkmark                 & 94.3          & 85.8           & 72.7              & 74.5              & 54.4              \\
    \midrule
         $\Delta$                   & 11.0          & 11.9           & \blue{+1.2}       & \blue{+1.6}       & \blue{+2.2}         \\
    \bottomrule
    \end{tabular}}    
    \label{tab:uniter-pretrain}
\end{table}

\begin{table}[t]
    \centering
    \caption{Pre-training gain from Vil-BERT.}
    \begin{tabular}{c|ccccc}
    \toprule
         \multirow{2}{*}{Pre-train} & \multicolumn{2}{c}{Image Retrieval}   & \multicolumn{3}{c}{VCR Validation}                        \\
                                    \cmidrule(lr){2-3}                                                \cmidrule(lr){4-6}
                                    & R@1       & R@5                       & Q$\rightarrow$A   & QA$\rightarrow$R  & Q$\rightarrow$AR  \\ 
    \midrule
         $\times$                   & 45.5      & 76.8                      & 69.3              & 71.0              & 49.5              \\
         \checkmark                 & 58.2      & 84.9                      & 72.4              & 74.5              & 54.0              \\
    \midrule
         $\Delta$                   & 12.7      & 8.1                       & \blue{+3.1}       & \blue{+3.5}       & \blue{+4.5}       \\
    \bottomrule
    \end{tabular}
    \label{tab:vilbert-pretrain}
\end{table}

\noindent \textbf{Optimization} -- For fine-tuning on VCR, the final output of the [CLS] token is utilized to predict whether the given answer or rationale is the correct choice. 
The VL Transformer is trained in an end-to-end fashion by minimizing the multi-class cross-entropy loss between the prediction for each response and the ground truth label. 
During inference, models are also comprehensively evaluated with the classification accuracy on the two sub-tasks, namely the holistic Q$\rightarrow$AR task (the accuracy set interaction of Q$\rightarrow$A and QA$\rightarrow$R).

Note that there are often two separable models for the two sub-tasks. 
The logical connection between these two is surprisingly ignored in the existing literature.
\section{Experimental Results and Findings}

Our findings are based on four popular SOTA VL-Transformers - UNITER~\cite{uniter}, ViLBERT~\cite{vilbert}, VL-BERT~\cite{VL-BERT} and VILLA~\cite{villa}.
We employed these four representative models for two reasons: 1) they cover a wide variety of different pre-training datasets, objectives, and architectures, and 2) the models are evaluated on VCR and the released codes can be rerun for reproduction.
After examining several failure cases in Figure~\ref{fig:case_study}, we found that these models often made mistakes on challenging queries. 
As the pretext tasks used by VL Transformers are mostly matching goal-driven rather than reasoning-oriented, we argue that they fail to do visual reasoning over commonsense scenes, which is the key to VCR. 
As a result, the superior performance is somewhat misleading. 
To explore this problem, we conducted extensive experiments and summarize some key findings below (Note that some experimental results have been moved to supplementary material due to space limitation.). 
This section elaborates on the following research questions:
\begin{itemize}
    \item \textbf{RQ1} -- How much does the pre-training help the downstream VCR task?
    \item \textbf{RQ2} -- Is the image modality really helpful for `visual reasoning' in VCR?
    \item \textbf{RQ3} -- Are there any correlations between the two separately trained models for the two sub-tasks? 
    \item \textbf{RQ4} -- Is it desirable to ignore the distinct connection between tags and objects?
\end{itemize}

\subsection{Limited Gain from Pre-training}\label{sec: pretraining}
Pre-training is of vital importance to downstream VL tasks, wherein the pretext training objectives play an important part. 
These pretext objectives, as detailed in Section~\ref{sec:pre-train-objective}, focus mainly on the conventional recognition ability.
Nevertheless, a good VCR model should be made cognition-aware and capable of visual reasoning. 
In light of this, the contribution of pre-training to VCR remains somewhat less trustworthy. 
As shown in Table~\ref{tab:villa-pretrain}, ~\ref{tab:uniter-pretrain} and ~\ref{tab:vilbert-pretrain}, the gains from pre-training for these matching-oriented tasks, \eg cross-modal retrieval~\cite{R3_TIP}, are quite substantial (more than 10 points in Table~\ref{tab:uniter-pretrain}). 
By contrast, when it comes to VCR, the performance gain is limited to $0\sim2$ points.
It illustrates that current VL Transformers benefit much more from the architecture itself than the pre-training objectives.

There are two possible explanations for this. First, these pretext objectives are specially designed for simple modality reconstruction and cross-modal matching. 
However, VCR demands fine-grained visual reasoning which cannot be achieved by the above means. 
Second, the domain shift between pre-training and VCR fine-tuning makes the transfer difficult. 
The images in the VCR dataset are about movie plots which are distinctive from the pre-training image captioning datasets~\cite{cc}.

\begin{figure*}[htbp]
  \centering
  \includegraphics[width=1.0\linewidth]{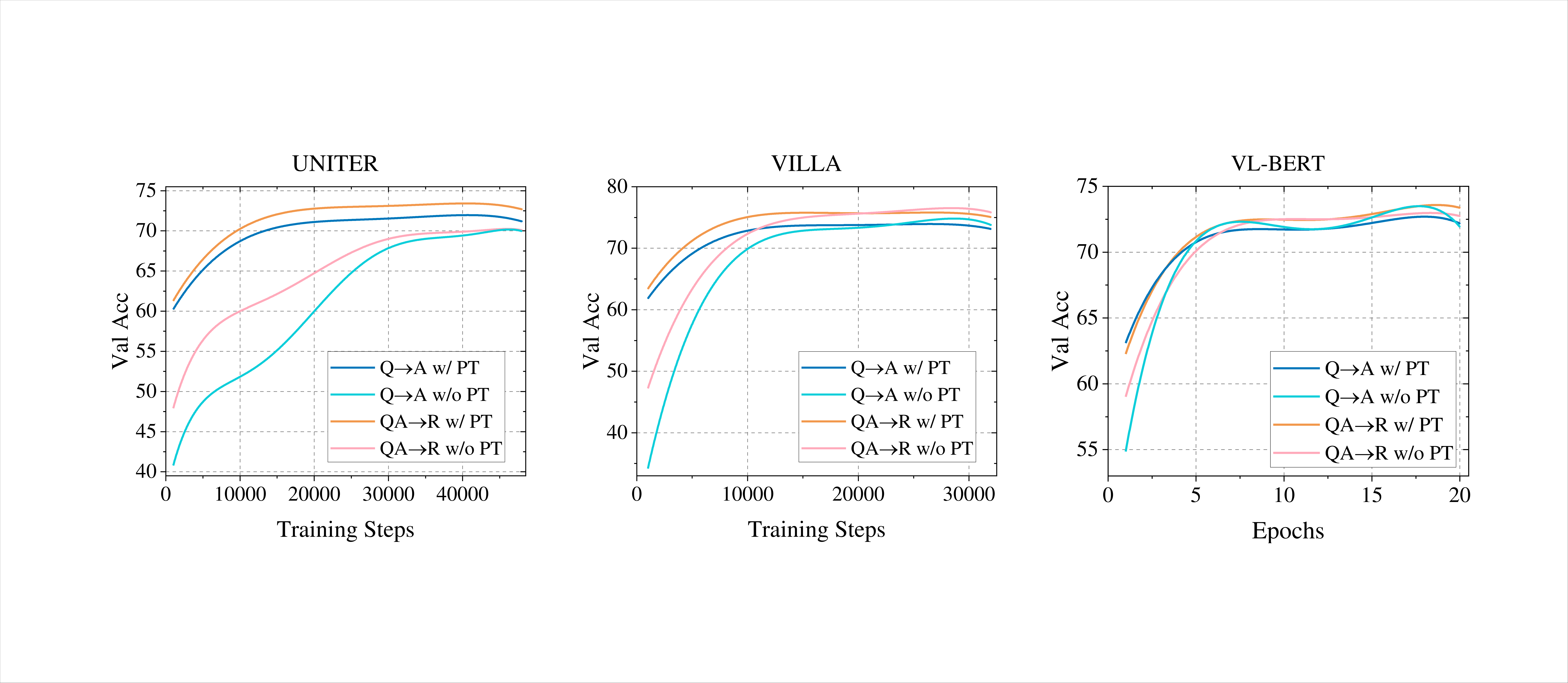}
  \caption{Convergence analysis of three VL Transformers with and without pre-training.}\label{fig:convergence_line}
\end{figure*}

We also investigated the convergence of three VL Transformers and display the results in Figure~\ref{fig:convergence_line}. 
It shows that pre-training boosts models with a good initialization point. 
With more training steps, the advantage from pre-training decays until the models with and without pre-training reach a similar level of performance. 

\subsection{Language Bias} \label{sec:language-bias}
It is well-known that VQA has been long affected by the language bias problem~\cite{lan-prior2,language-bias1}. 
It refers to the correlation shortcut between textual questions and answers. 
To study whether VCR is afflicted with this problem, we tested two settings: VL Transformers respectively removing query and image  (see Figure~\ref{fig:pipeline}), and observed the results. 
Table~\ref{tab:comparison_image_query} shows that for Q$\rightarrow$A, the model variants without query and without image lead to similar performance degradation. 
However, for the sub-task of QA$\rightarrow$R, there exists a significant performance gap between these two variants, \ie the difference between the former variant and the full model is over 2$\times$ more than that of the latter. 
It is mainly because of the language shortcut between textual inputs (One such example can be seen in Figure~\ref{fig:teaser}, that only the correct r1 contains the [Person3] tag.). 
VL Transformers tend to utilize such bias for prediction rather than performing visual reasoning. 

\begin{figure}[!t]
  \centering
  \includegraphics[width=1.0\linewidth]{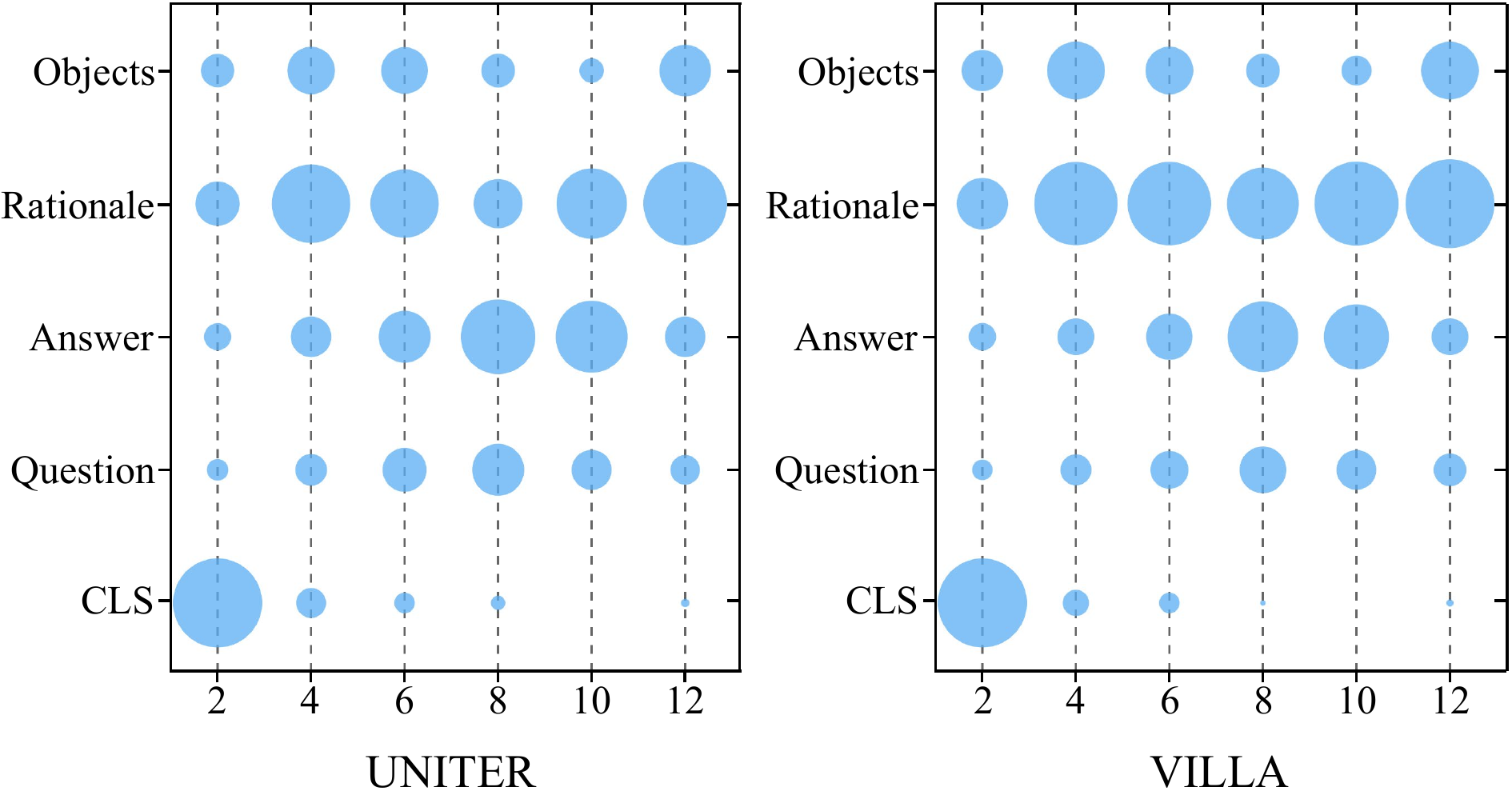}
  \caption{Attention distribution from the token of [CLS]. We empirically selected even layers for demonstration.}\label{fig:attention}
\end{figure}

To further validate this hypothesis, we then explored the attention weight distribution of these models.
In general, the attention weights express how much other elements contribute to the learning of the current element. 
We first estimated the learned attention weights from UNITER~\cite{uniter} and VILLA~\cite{villa} based upon the [CLS] token, as the output from [CLS] is leveraged for predicting the correct response.
It can be observed in Figure~\ref{fig:attention} that the two models pay too much attention to the textual rationale and answer elements (except for the [CLS] token itself in the second layer) while focusing less on the visual objects. 
Since the goal of VCR is to pursue reasoning with images, models making decisions without the involvement of vision seem less desirable. 

\begin{figure}[htbp]
  \centering
  \includegraphics[width=1.0\linewidth]{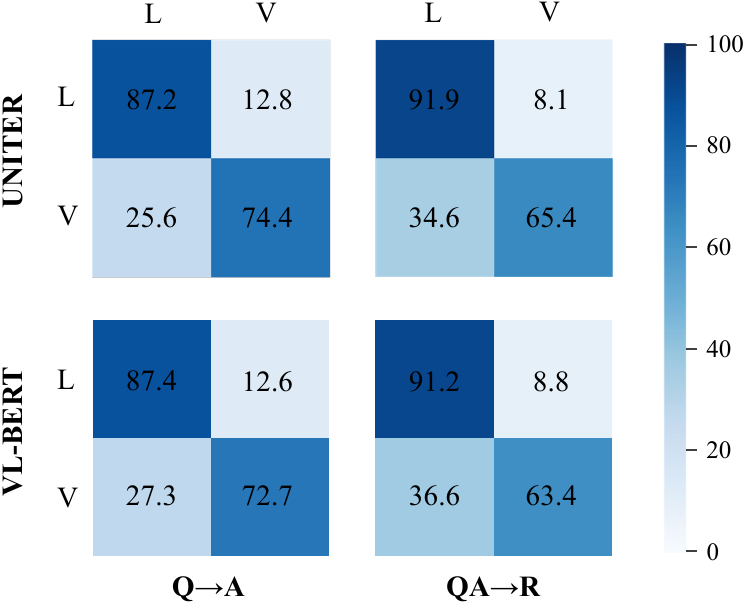}
  \caption{The attention weight distribution from the modality of each row to the modalities of columns.}\label{fig:confusion}
\end{figure}

\begin{figure}[htbp]
  \centering
  \includegraphics[width=1.0\linewidth]{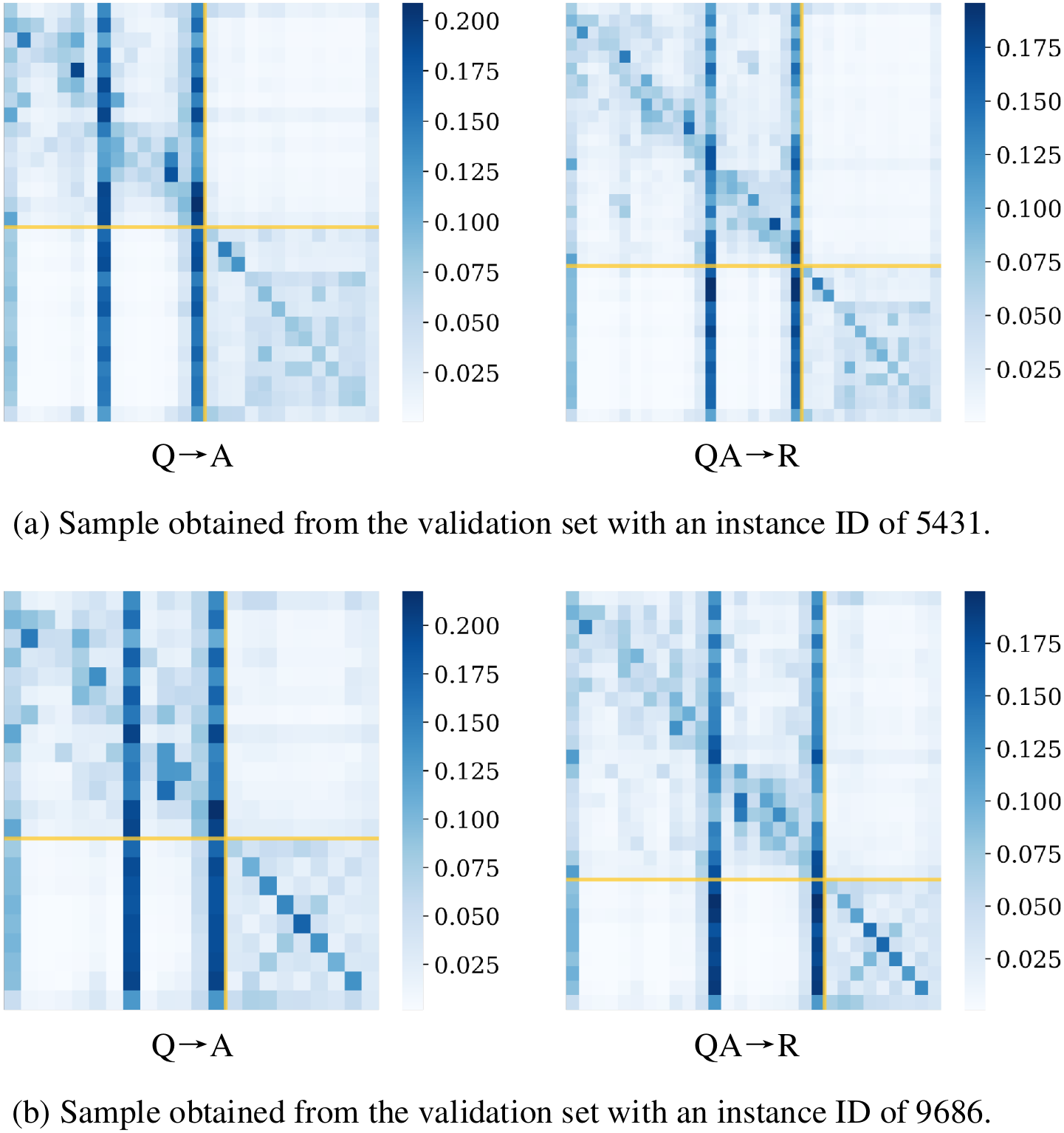}
  \caption{Qualitative results of the attention map from the last self-attention layer in VL-BERT. Each row represents the attention weight of a given input token with respect to all tokens. We use the yellow lines to separate textual tokens from visual tokens.}\label{fig:attention_map}
\end{figure}

Thereafter, we further evaluated the attention weights from one modality to the other, and illustrate the results in Figure~\ref{fig:confusion}. 
We can observe that the language modality sees almost only itself, especially for the QA$\rightarrow$R sub-task. 
It further validates that the language bias dominates the prediction of both sub-tasks, and QA$\rightarrow$R is affected more. 
By contrast, the vision modality looks more balanced with respect to its attention distribution. 
This modality bias issue is further reflected by two examples in Figure~\ref{fig:attention_map}, where it can be seen that each modality mainly focuses on its own encoded tokens.

\begin{table}[!t]
    \centering
    \caption{Performance of three VL Transformers within three variants: full model, without query and without image. Note that  Q$\rightarrow$AR tells the interaction between correctly predicted Q$\rightarrow$A and QA$\rightarrow$R instances.}
    \scalebox{1}{
    \begin{tabular}{l|ccc}
    \toprule
         Model                    & Q$\rightarrow$A                 & QA$\rightarrow$R                & Q$\rightarrow$AR\\
    \midrule
         UNITER                   & 74.4                            & 76.9                            & 57.5 \\
         \emph{\quad w/o} query   & 59.3 (\textcolor{blue}{-15.1})  & 57.2 (\textcolor{blue}{-19.7})  & 34.5 (\textcolor{blue}{-23.0}) \\
         \emph{\quad w/o} image   & 59.6 (\textcolor{blue}{-14.8})  & 68.6 (\textcolor{red}{-8.3})    & 41.0 (\textcolor{blue}{-16.5}) \\
    \midrule
         VL-BERT                  & 72.6                           & 74.0                             & 54.0  \\
         \emph{\quad w/o} query   & 56.4 (\textcolor{blue}{-16.2})  & 53.5 (\textcolor{blue}{-20.5})  & 30.8 (\textcolor{blue}{-23.2}) \\
         \emph{\quad w/o} image   & 58.8 (\textcolor{blue}{-13.8})  & 66.0 (\textcolor{red}{-8.0})    & 38.9 (\textcolor{blue}{-15.1}) \\
    \midrule
         VILLA                    & 75.4                           & 78.7                             & 59.5  \\
         \emph{\quad w/o} query   & 60.6 (\textcolor{blue}{-14.8})  & 58.8 (\textcolor{blue}{-19.9})    & 36.2 (\textcolor{blue}{-23.3}) \\
         \emph{\quad w/o} image   & 60.5 (\textcolor{blue}{-14.9})  & 71.0 (\textcolor{red}{-7.7})      & 43.1 (\textcolor{blue}{-16.4}) \\
    \bottomrule
    \end{tabular}}
    \label{tab:comparison_image_query}
\end{table}


\subsection{Sparse Correlation between Two Models}
\begin{figure}[!t]
  \centering
  \includegraphics[width=1.0\linewidth]{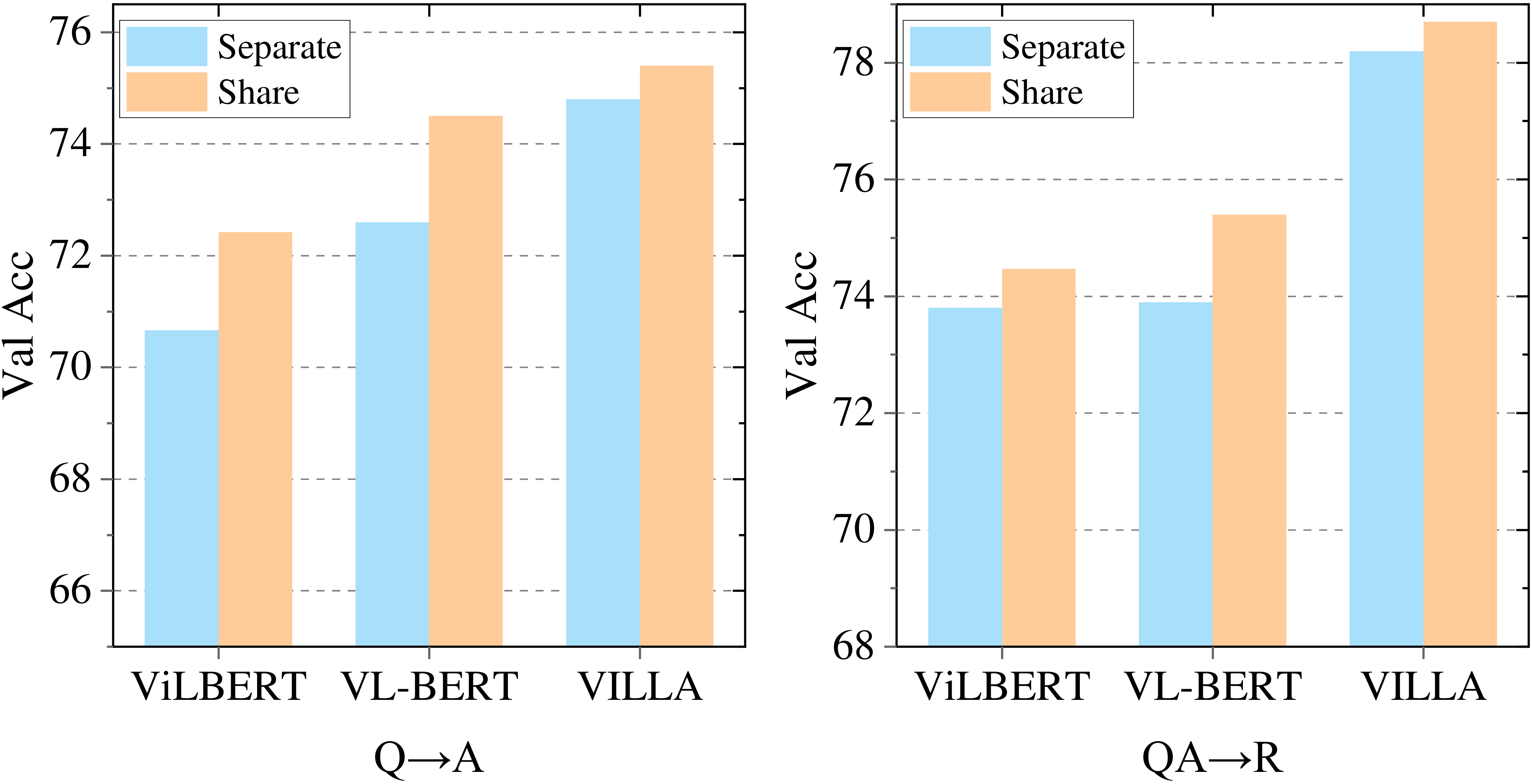}
  \caption{Performance comparison of three models with sharing and separate parameters.}\label{fig:share_param}
\end{figure}

Figure~\ref{fig:pipeline} shows that VL Transformers take the Q$\rightarrow$A and QA$\rightarrow$R as two independent processes. 
In other words, there are two separate models with similar architectures and training protocols. 
However, separately treating Q$\rightarrow$A and QA$\rightarrow$R deteriorates the visual scene understanding, considering that these two processes share a common goal~\cite{tip, vcr_align}.
In the following, we investigated the correlation between these two models from two angles.

\begin{figure*}[!t]
  \centering
  \includegraphics[width=1.0\linewidth]{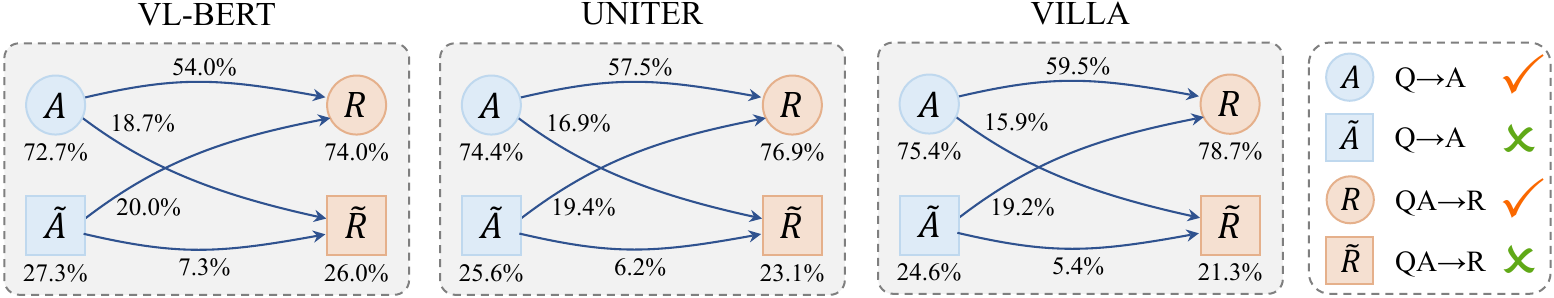}
  \caption{Proportion of correctly predicted instances from three VL Transformers and their intersection of four states in VCR.}\label{fig:intersection}
\end{figure*}

One intuitive idea is to study the overlapped instances between Q$\rightarrow$A and QA$\rightarrow$R, see Figure~\ref{fig:intersection}.
Note that the query to QA$\rightarrow$R is the given question concatenated with the right answer. 
Given the same question, on the one hand, we observed that around 3/4 of correctly predicted answers lead to the right rationale (from round box A to round box R).
This shows that some answers are at least not predicted based on the same reasoning as humans. 
On the other hand, we found that a large proportion of wrongly predicted answers (3/4) correspond to the right rationale (from the square box A to round box R). 
We suspect that one possible reason is due to the shortcut learning between correct answers and rationales, as discussed in Section~\ref{sec:language-bias} because the input to QA$\rightarrow$R contains the ground-truth answer instead of the predicted one.

As discussed before, existing methods all employ two models to separately tackle Q$\rightarrow$A and QA$\rightarrow$R. 
This makes us wonder, are there any differences between these two models? 
Or does the separation of the two sub-tasks really allow the two models to better deal with answering and reasoning? 
To answer this question, we conducted tests with the same model for both sub-tasks. 
Figure~\ref{fig:share_param} illustrates that when employing the same model for these two, the model performance is slightly increased.
We speculate that the cause of this phenomenon is the doubling of the dataset employed for training. 
This experiment, on the other hand, shows that VL Transformers do not differentiate these two sub-tasks, despite the fact that the latter one requires more visual commonsense. 

\begin{table}[!t]
    \centering
    \caption{Performance of three VL Transformers within three variants: full model, without tags, and with the random replacement of tags.}
    \scalebox{1}{
    \begin{tabular}{l|c|ccc}
    \toprule
         Model                    & Q$\rightarrow$A                 & QA$\rightarrow$R                & Q$\rightarrow$AR\\
    \midrule
         UNITER                   & 74.4                            & 76.9                            & 57.5 \\
         \emph{\quad w/o} tag     & 68.6 (\textcolor{blue}{-5.8})   & 72.8 (\textcolor{blue}{-4.1})   & 50.5 (\textcolor{blue}{-7.0}) \\
         \emph{\quad repl.} tag   & 73.5 (\textcolor{blue}{-0.9})   & 76.3 (\textcolor{blue}{-0.6})   & 56.4 (\textcolor{blue}{-1.1}) \\
    \midrule
         VL-BERT                  & 72.6                            & 74.0                            & 54.0  \\
         \emph{\quad w/o} tag     & 65.7 (\textcolor{blue}{-6.9})   & 69.5 (\textcolor{blue}{-4.5})   & 46.1 (\textcolor{blue}{-7.9}) \\
         \emph{\quad repl.} tag   & 72.5 (\textcolor{blue}{-0.1})   & 73.9 (\textcolor{blue}{-0.1})   & 53.9 (\textcolor{blue}{-0.1}) \\
    \midrule
         VILLA                    & 75.4                            & 78.7                            & 59.5  \\
         \emph{\quad w/o} tag     & 69.7 (\textcolor{blue}{-5.7})   & 74.8 (\textcolor{blue}{-3.9})   & 52.5 (\textcolor{blue}{-7.0}) \\
         \emph{\quad repl.} tag   & 74.5 (\textcolor{blue}{-0.9})   & 78.2 (\textcolor{blue}{-0.5})   & 58.5 (\textcolor{blue}{-1.0}) \\
    \bottomrule
    \end{tabular}}
    \label{tab:tag}
\end{table}

\subsection{Incompleteness of Tag Handling}
\begin{figure*}[!t]
  \centering
  \includegraphics[width=1.0\linewidth]{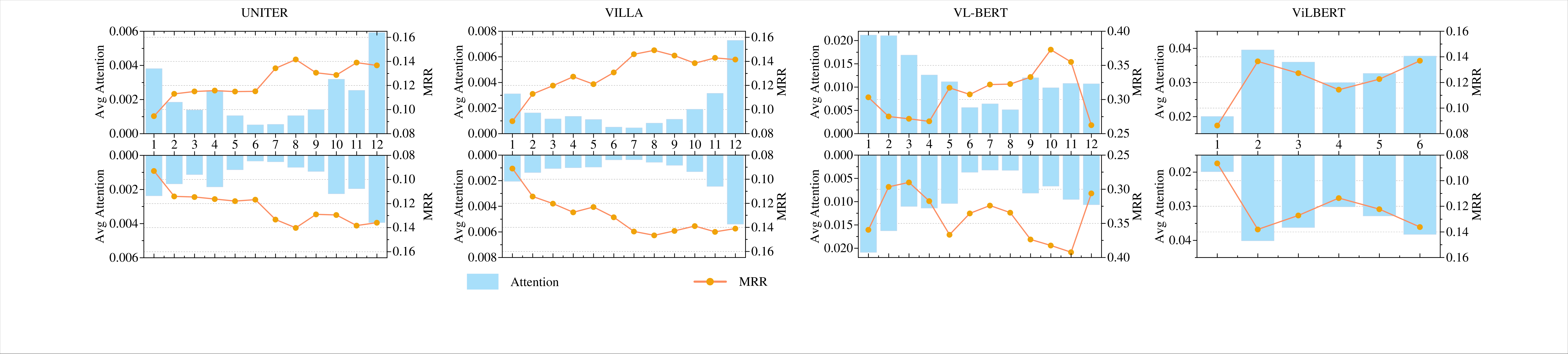}
  \caption{Average attention weights and MRR value with respect to each tag for different VL Transformer layers (1 to 12). We use three typical VL Transformers for demonstration. Top: Q$\rightarrow$A; Bottom: QA$\rightarrow$R.}\label{fig:mrr}
\end{figure*}
In VCR, the questions, answers and rationales are written in a mixture of rich natural language as well as detection tags, like `[person1 \raisebox{-0.1cm}{\includegraphics[width=0.05\linewidth]{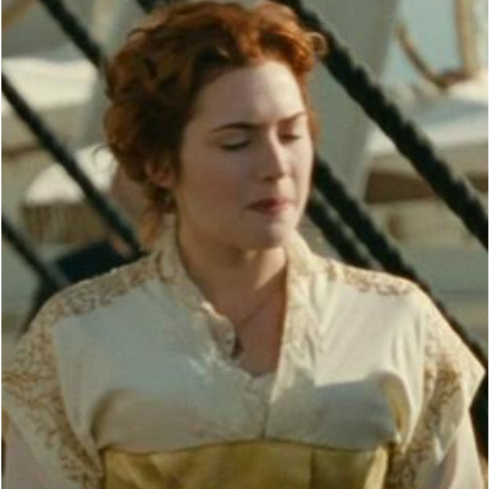}}]': this helps to provide a direct link between the textual description of an object and its corresponding image region. 
Most VL Transformers, however, ignore this information for modeling simplicity. 
The typical handling of the tag label is to directly utilize the text input, \eg `[person1 \raisebox{-0.1cm}{\includegraphics[width=0.05\linewidth]{img/person3.pdf}}]'$\Rightarrow$person (a neutral name). 

To investigate the importance of these tag labels, we first removed the tag input and observed the results in Table~\ref{tab:tag}. 
It can be seen that the model performance degrades to some extent.
The key reason is that the input sentences lack important subjects, and therefore makes VL Transformers confusing.
We then randomly replaced each tag with another one, \eg `[table]' $\Rightarrow$ `[person]'. 
The results in Table~\ref{tab:tag} demonstrate that the models show only minor deterioration. 
This illustrates that existing handling for tags is largely limited. 
That is, randomly replacing tags barely impacts the performance though the link between each tag and object is deliberately broken.

Like in Section~\ref{sec:language-bias}, we also studied the attention distribution, especially the attention weights attached to each given tag.
In particular, we used two metrics to quantify this effect: 

\textbf{Avg Attention} is adopted to count the averaged attention values between the given tag and the visual object to which it referred (upper bound is 1.0),
\begin{equation}
    \label{eq: Avg_Attention}
    att_l = \frac{1}{\sum_{i=1}^{n}m_i}\sum_{i=1}^{n}\sum_{j=1}^{m_i}score(t_{ij}, o_{ij}),
\end{equation}
where $score(t_{ij}, o_{ij})$ denotes the attention score from the tag token $t_{ij}$ to its corresponding visual object $o_{ij}$ in layer $l$, $n$ and $m_i$ are the number of samples in the validation set and the number of tags contained in each sample, respectively.

\textbf{MRR} is employed to estimate the rank of the true object based on the predicted attention values for each tag (upper bound is 1.0),
\begin{equation}
    \label{eq: MRR}
    mrr_l = \frac{1}{\sum_{i=1}^{n}m_i}\sum_{i=1}^{n}\sum_{j=1}^{m_i}\frac{1}{rank(t_{ij}, o_{ij})},
\end{equation}
where $rank(t_{ij}, o_{ij})$ is the ranking of the $score(t_{ij}, o_{ij})$ among the attention scores between $t_{ij}$ and all objects (5 to 13 per image).
As shown in Figure~\ref{fig:mrr}, for all three VL Transformers, both values are pretty low in relation to their upper bound (1.0). This result demonstrates that the  link between the tag and its attached object is almost nonexistent. As a result, whether the VL Transformers perform reasoning remains doubtful since such an important correlation is ignored. 
\section{What to Do Next?}
The above findings tell us that existing VL Transformers do not offer a good solution for visual commonsense understanding. 
As good as the recognition capability is, a VCR model is expected to be endowed with more reasoning strengths. 
In what follows, we outline several possible directions worth exploring in the next:

\textbf{Dataset} -- 
Curating more challenging datasets which can circumvent the shortcut modeling problem encountered by existing VL Transformers. 
In addition, probing other means of visual commonsense understanding instead of giving explanations to question answering is also of great potential, \eg the collection of spatial or attribute commonsense.

\textbf{Evaluation Metric} -- 
Designing specific metrics for quantifying the reasoning capability of models, for which the current vanilla accuracy metric is limited in its form.
Besides time-intensive subjective evaluations, we can also gain insights from the text generation tasks like machine translation~\cite{NMT} and image captioning~\cite{image_caption4}. 

\textbf{Pre-training Task} --
Making the pre-training objectives focus more on cognition and understanding.
In particular, how to enhance VL Transformers with reasoning strengths is an interesting path for improving downstream VCR. 
One possible attempt is to incorporate large-scale knowledge into pre-training pretext tasks.

\textbf{De-biasing} -- 
Guiding models with suitable de-biasing tricks can help overcome the language bias problem.
It is worth noting that the bias problem is more challenging than VQA because of the spurious correlation between a single query and response. 
In contrast to VCR, the bias in the sister VQA domain results from the statistical shortcut between question type and answers\footnote{The question type is often referred to the first few words of the given question, \eg how many. Answers in VQA datasets are mostly composed of a few keywords.}, whereby some trivial loss re-balancing tricks can be employed~\cite{lan-prior1}.

\textbf{Model Architecture} -- 
Developing suitable model architectures to take advantage of the unique tag-object label and the two-step visual commonsense understanding. 
For instance, we can approach Q$\rightarrow$A and QA$\rightarrow$R simultaneously with collaboration, where the common image information provides an essential proxy to achieve this goal. 

\textbf{Prompting on Large Language Models} -- 
One last promising future direction is to prompt the pre-trained large language models, such as ChatGPT\footnote{https://chat.openai.com/}.
A typical approach is to decode the image into text, and then the generalization capability of these large models can be leveraged to provide the correct explanation for visual questions.
Nevertheless, the captioning quality of another proxy model is still under questioning.
\section{Related Work}
\subsection{Vision-Language Transformers}
The success of Transformers in the field of Natural Language Processing (NLP)~\cite{bert} and Computer Vision (CV)~\cite{ViT,DETR,SegFormer,ViLT} brings a lot of progress for multi-modal vision and language tasks~\cite{VL_Transformers1, videos,R1-4_2,FLAVA,R3_5_ref2}.
Based on how the vision and language branches are fused, current VL Transformers can be roughly categorized into single-stream  (\eg UNIMO~\cite{unimo} and SOHO~\cite{SOHO}) and dual-stream cross-modal Transformers (\eg LXMERT~\cite{lxmert} and ALBEF~\cite{ALBEF}). 
A typical VL Transformer model often employs the \emph{pretrain-then-finetune} learning schema: the model is first pre-trained on large vision-text datasets and then fine-tuned on downstream tasks by transferring their rich representations from pre-training. 
Specifically, the pretext tasks play a vital role in pre-training, where masked language modeling, masked region prediction, and image-text matching are extensively studied. 
The fine-tuning step mirrors that of the BERT model~\cite{bert}, which includes a task-specific input, output, and objective. The pre-trained model is thereafter optimized to maximize the performance on the corresponding vision-and-language task.  
The VL Transformers mainly help in the following three groups of downstream tasks: cross-modal matching, cross-modal reasoning, and vision language generation~\cite{survey_of_VLTransformer}. 
The first group focuses on learning cross-modal correspondences between vision and language, such as image text retrieval and visual referring expression. 
Reasoning ones require VL Transformers to perform language reasoning based on visual scenes, such as VQA. 
The last group aims to generate the targets of one modality given the other as input~\cite{generation,lxmert}. The desired visual or textual tokens are decoded in an auto-regressive generation manner.

\subsection{Visual Commonsense Reasoning}
Recently, multimodal tasks have garnered increasing research attention~\cite{IMP_GCN, MAML, DMRL, micro-video, sentiment, video_retrieval, video_retrieval2, LQN_1, LQN_2}, with a notable example being Visual Question Answering (VQA). However, conventional VQA models often face limitations due to their black-box reasoning capabilities.
To move one step further, VCR was presented to supplement VQA by inferring the rationale behind the question-answering process.
Some initial work designs specific architectures to address VCR for the purpose of finer-grained visual understanding.
For instance, R2C~\cite{r2c} performs three inference steps - grounding, contextualization, and reasoning, to move towards cognition-level understanding step by step. 
Inspired by the neuron connectivity of the human brain, CCN~\cite{CCN} designs a connective cognition network to globally and dynamically integrate the local visual neuron connectivity. 
For explicit cross-modal representation learning, syntactic information is incorporated into the visual reasoning and natural language understanding~\cite{TMM_zhangxi}.
Recent progress on the VCR leaderboard is mostly derived from VL Transformers. 
For example, UNITER~\cite{uniter} utilizes a single-stream encoder and four elaborate pre-training tasks to learn universal image-text representations for various downstream multi-modal tasks.
ViLBERT~\cite{vilbert} employs a dual-stream fusion encoder with co-attention layers to model the inter-modality interactions.
MERLOT~\cite{merlot} first learns commonsense representations of multi-modal events by pre-training over millions of videos and then transfers them to the target images in the VCR dataset. 
Despite the impressive performance achieved by these VL Transformer models, whether they truly possess visual commonsense remains an intriguing yet under-explored question.
\section{Conclusion}
This paper recognizes several limitations of utilizing existing VL Transformers to VCR, which takes an essential step towards visual commonsense understanding. 
Though the results on the benchmark dataset are impressive, we argue that these numbers are less trustworthy in terms of visual reasoning -- the key ingredient of VCR. 
On the flip side, this `all in Transformer' wave may mislead the VCR research in the wrong direction, preventing the community from pursuing models that are reasoning-aware.
To the best of our knowledge, we are the first to comprehensively study this problem in literature though we strongly believe we are not the only one afflicted with it.
With the discovery of this problem and the tentative proposal for several future directions, we hope this work can help motivate more interesting and plausible ideas for visual commonsense reasoning in the future.

\begin{acks}
This research is supported by the National Research Foundation, Singapore under its Strategic Capability Research Centres Funding Initiative, and the National Natural Science Foundation of China, No.:U1936203. Any opinions, findings and conclusions or recommendations expressed in this material are those of the author(s) and do not reflect the views of National Research Foundation, Singapore.
\end{acks}

\bibliographystyle{format/ACM-Reference-Format}
\balance
\bibliography{VCR-Retro}

\end{document}